\documentclass[10pt,twocolumn,letterpaper]{article}

\usepackage{graphicx}
\usepackage{mathdefs}
\usepackage{gfxmlinverse}
\usepackage{booktabs} 
\usepackage{caption}
\usepackage{color}
\usepackage{nicefrac}
\usepackage{units}
\usepackage{cvpr}
\usepackage{times}
\usepackage{epsfig}
\usepackage{amsmath}
\usepackage{amssymb}
\usepackage{multirow}
\usepackage{tabularx}

\graphicspath{{figures/}}

\usepackage[breaklinks=true,bookmarks=false]{hyperref}

\cvprfinalcopy 

\def\cvprPaperID{****} 

\begin{document}

\title{Inverse Transport Networks}

\author{Chengqian Che\\
Carnegie Mellon University\\
\and
Fujun Luan\\
Cornell University\\
\and
Shuang Zhao\\
University of California, Irvine\\
\and
Kavita Bala\\
Cornell University\\
\and
Ioannis Gkioulekas\\
Carnegie Mellon University
}

\maketitle

\begin{abstract}
We introduce inverse transport networks as a learning architecture for inverse rendering problems where, given input image measurements, we seek to infer physical scene parameters such as shape, material, and illumination. During training, these networks are evaluated not only in terms of how close they can predict groundtruth parameters, but also in terms of whether the parameters they produce can be used, together with physically-accurate graphics renderers, to reproduce the input image measurements. To enable training of inverse transport networks using stochastic gradient descent, we additionally create a general-purpose, physically-accurate differentiable renderer, which can be used to estimate derivatives of images with respect to arbitrary physical scene parameters. Our experiments demonstrate that inverse transport networks can be trained efficiently using differentiable rendering, and that they generalize to scenes with completely unseen geometry and illumination better than networks trained without appearance-matching regularization.
\end{abstract}

\maketitle

\section{Introduction}\label{sec:intro}

Acquiring models of the world has been a long-standing challenge in computer graphics and vision.
During the past four decades, a variety of physics-based measurement systems~\cite{marschner1998inverse,terzopoulos1989physically,ramamoorthi2001signal,patow2005survey,patow2003survey,bertero1988ill,weyrich2009principles} have been devised.
These systems aim to reconstruct the shape and material of real-world objects under controlled or uncontrolled illumination and from images captured under a variety of radiometric devices (e.g., projector-camera systems, time-of flight devices). 
Following the image acquisition stage, these systems utilize an inference algorithm to solve for the physical parameters of interest, for example, shape or material.
We can broadly classify such inferential algorithms into physics-based and learning-based techniques.

Physics-based techniques infer unknowns by optimizing for parameter values until they can be used, together with a physics-based image formation model, to reproduce input images.
In certain cases, this optimization can be performed analytically (e.g., photometric stereo~\cite{ikeuchi1981determining}, direct reflectometry~\cite{marschner1998inverse}).
More generally, this optimization boils down to seeking for parameter values with which forward simulations best resemble the measurements.
This framework is usually termed \emph{analysis-by-synthesis} in computer vision or \emph{inverse rendering} in graphics~\cite{loper2014opendr,Gkioulekas2013:IVR,Zhao:2016:DSP,Khungurn:2015:MRF,gkioulekas2016inverse,levis2015clouds}. 
By accurately modeling the underlying physics of light transport, these techniques can produce high-fidelity estimates for fully general inputs, at the cost of high, and often prohibitive, computational requirements. 

Learning-based techniques, on the other hand, use supervised and unsupervised data to create functions that approximately map sensor measurements directly to physical parameters.
In the last few years, predominantly these functions take the form of multi-layer neural networks, which can be efficiently optimized on large training datasets using stochastic optimization~\cite{li2017modeling,rematas2016deep,tang2012deep,zhou2015learning,narihira2015direct,wang2015designing,li2015depth}.
These techniques allow for efficient inference, but do not offer guarantees about the physical plausibility and interpretability of predicted parameter values.
Additionally, since these methods are purely data-driven and do not attempt to reproduce the underlying physics, they often generalize poorly to inputs that are underrepresented in the training dataset.

We seek to combine the complementary advantages of the physics- and learning-based approaches for the inference of physical scene properties, by proposing a new physics-aware learning technique 
that we term \emph{inverse transport networks}.
Taking inspiration from recent work on combining physics and learning~\cite{lime,shrenderer,liu2017material,mofa,neuralmeshrenderer}, inverse transport networks are trained by solving a regularized optimization problem that forces the network to produce output parameters that not only match ground-truth values but also reproduce the input images (when used as input to a forward physics-based renderer).
Previously, this kind of regularized training could only be used in cases where the forward physics were sufficiently simple (i.e., can be differentiated analytically), severely restricting its applicability.
By contrast, our inverse transport networks can be used with arbitrarily complex forward physics that can capture global illumination effects such as interreflections and subsurface scattering.
To this end, we introduce a new forward simulation engine that can be used for \emph{efficient}, \emph{general-purpose} and \emph{physically accurate} Monte Carlo differentiable rendering. Finally, to demonstrate the general-purpose nature and improved predictive performance of inverse transport networks, we evaluate them on the task of homogeneous inverse scattering, an inverse problem involving highly multi-path and multi-bounce light transport.

\section{Related work}

\boldstart{Analysis-by-synthesis in physics-based vision.} Physics-based algorithms for recovering scene parameters conceptually comprise three steps: (i)~formulate an approximate image formation (or forward rendering) model as a function of the scene parameters; (ii)~analytically
derive an expression for the derivative of the forward model with respect to
those parameters; (iii)~use gradient-based optimization to solve an
analysis-by-synthesis objective comparing measured and synthesized
images. This approach has been
used to recover shape~\cite{gargallo2007minimizing,solem2005geometric,delaunoy2011gradient},
material~\cite{romeiro2010blind,mukaigawa2010analysis,nishino2009directional}, and illumination~\cite{mei2011illumination}, either independently of each other or jointly~\cite{barron2015shape,yoon2010joint,lombardi2012reflectance,lombardi2016reflectance,oxholm2014multiview}.

\boldstart{Differentiable rendering.} Limiting the applicability of this general approach is the need to formulate a new forward model, as well as the need to analytically compute its derivatives, specifically for each reconstruction problem.
Differentiable renderers such as OpenDR~\cite{loper2014opendr} have been proposed as a means to remove this obstacle, by providing a general-purpose 
framework that can be differentiated with respect to shape, reflectance, and illumination parameters.
To ensure analytical differentiability, all of the above approaches use \emph{approximate} forward models, most often by ignoring complex light transport effects such as inter-reflections and subsurface scattering. This makes these methods inapplicable to situations where these effects are dominant. Some solutions have been developed for the problem of inverse scattering~\cite{Gkioulekas2013:IVR,Zhao:2016:DSP,Khungurn:2015:MRF,gkioulekas2016inverse,levis2015clouds}, which use physically accurate, but limited to very specialized light transport simulations, Monte Carlo differentiable renderers.

\boldstart{Combining deep learning with rendering.} Recently, a number of works have emerged that propose using renderers not for analysis-by-synthesis, but as parts of learning architectures. The most popular approach is to replace the decoder network in an auto-encoder pipeline~\cite{vincent2010stacked,kingma2013auto} with a rendering layer that takes as input the parameters predicted by the encoder and produces as output synthesized images. This encoder-renderer architecture was first proposed by Wu et al.~\cite{wuneural}, who used a non-photorealistic renderer to achieve categorical interpretability (e.g., ``a boy and girl stand
next to a bench''). Similar architectures have subsequently been proposed for physics-based inference of parameters such as surface normals, illumination, and reflectance~\cite{lime,shrenderer,liu2017material,mofa,neuralmeshrenderer}. Alternatively, renderers have been incorporated into adversarial learning pipelines for image-to-image translation tasks~\cite{tung2017adversarial}.

\section{Our Approach}
\label{sec:approach}

\boldstart{Problem setting.} We are interested in inverse problems where the unknowns are physical properties (e.g., geometry, material (optical) parameters, and illumination) of a scene imaged by a radiometric sensor.
Each image captured by the sensor records photons interacting multiple times with the surfaces and interior of objects in the scene, in a way that depends on the scene parameters.
We will refer to this complex \emph{light transport} process using $\opT\paren{\bpi}$, where $\bpi$ are the relevant physical parameters of the scene: $\bpi = \curly{\text{sensor}, \text{geometry}, \text{material}, \text{illumination}}$.

A long-standing approach for solving such inverse problems in computer vision and graphics is \emph{analysis by synthesis}, also known as \emph{inverse rendering}.
Given image measurements $I$, we search for parameters $\bpi$ that, when used to synthesize images, can closely match the measurements:
\begin{equation}
\label{eq:inverse:rendering}
\hat{\bpi} = \argmin_{\bpi} \norm{I - \opT\paren{\bpi}}^2.
\end{equation}
Analysis by synthesis offers two key advantages that make it an attractive algorithm for solving inverse problems of this kind.
First, it is a general-purpose procedure that can be applied to arbitrary scenes.
This is thanks to the advent in computer graphics of \emph{forward rendering} algorithms that can accurately simulate light transport $\opT\paren{\bpi}$ of arbitrary complexity, including global illumination effects such as interreflections, specular and refractive caustics, and multiple scattering.
Second, it is often possible to derive guarantees about the fidelity of the reconstructed parameters $\hat{\bpi}$, through analysis of the underlying physics.
Unfortunately, solving optimization problem~\eqref{eq:inverse:rendering} is often a computationally intensive process: even when it is possible to compute derivatives $\partial \opT\paren{\bpi} / \partial \bpi$ for gradient descent optimization, finding a (local) minimum of Equation~\eqref{eq:inverse:rendering} requires performing thousands of expensive rendering operations.

An alternative methodology for solving such inverse problems is through supervised learning.
Given a training set of image measurements $\curly{I_d}$ and groundtruth parameters $\curly{\bpi_d}$, $d = 1, \ldots, D$, we first use empirical risk minimization to train a parametric regression model $\opN\bracket{\bw}$, e.g., a neural network, that directly maps images to parameters:
\begin{equation}
\hat{\bw} = \argmin_{\bw} \sum_{d=1}^D \norm{\bpi_d - \opN\bracket{\bw}\paren{I_d}}^2.\label{eq:erm}
\end{equation}
Once the trained network $\opN\bracket{\hat{\bw}}$ is available, we can use it to efficiently obtain estimates $\hat{\bpi}$ for the physical parameters underlying new image measurements $I$, by performing a \emph{forward pass} operation: $\hat{\bpi} = \opN\bracket{\hat{\bw}}\paren{I}$.
This efficiency comes with the caveat that it is difficult to obtain guarantees about the quality of the estimates $\hat{\bpi}$. This becomes particularly important when we use the network to process images of scenes that are not well represented in the training set, e.g., very different shapes or illumination. Given the highly nonlinear mapping $\opT$ from scene to images, it can be challenging for networks to generalize to unseen scenes.

\subsection{Inverse Transport Networks}
\label{ssec:itn}
\begin{figure*}[t]
	\centering
	\includegraphics[width=\textwidth]{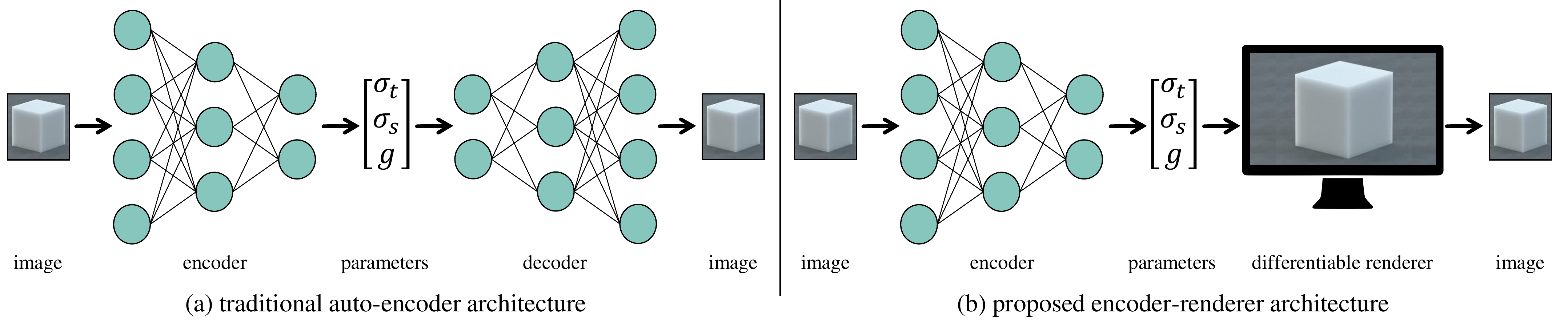}
	\caption{\label{fig:architecture}
		\textbf{Inverse transport networks:}
		(a)~Traditional autoencoder network architectures use two neural networks, encoder and decoder, to learn intermediate representations of input images that are subsequently remapped into images. (b)~Our proposed inverse transport network architecture replaces the decoder with a differentiable physics-based renderer, which improves generalization performance by acting as a regularization term on the encoder.}
\end{figure*}

We seek to combine the efficiency of learning with the generality of analysis by synthesis. For this, we propose to regularize the loss function~\eqref{eq:erm} used to train a regressor network with a term that closely resembles the loss function~\eqref{eq:inverse:rendering} optimized by analysis by synthesis:
\begin{equation}
\label{eq:erm:regularized}
\begin{aligned}
\hat{\bw} =& \argmin_{\bw} \sum_{d=1}^D \bigg[
\underbrace{\norm{\bpi_d - \opN\bracket{\bw}\paren{I_d}}^2}_{\text{supervised loss}}\\
&+ \lambda \underbrace{\norm{I_d - \opT\paren{\opN\bracket{\bw}\paren{I_d}}}^2}_{\text{regularization}} \bigg].
\end{aligned}
\end{equation}
The regularization term in Equation~\eqref{eq:erm:regularized} forces the neural network to predict parameters $\bpi_d$ that not only match the groundtruth, but also can reproduce the input images when used as input to forward rendering.
This has two desirable effects as follows.
First, the parameters predicted by the network are likely to be close to those that would be obtained from analysis-by-synthesis or inverse rendering, as the regularization term in Equation~\eqref{eq:erm:regularized} is equivalent to the analysis-by-synthesis loss~\eqref{eq:inverse:rendering}.
Second, the regularization term forces the regression function $\opN\bracket{\hat{\bw}}$ implemented by the neural network to be approximately equal to the inverse of the light transport operator $\opT$, that is, $\opN\bracket{\hat{\bw}}\approx \opT^{-1}$.
Given that $\opT$ models the physics of light transport that apply generally to all possible scenes, we expect the resulting neural network generalize well to novel scenes.
Accordingly, we term networks trained with the loss function~\eqref{eq:erm:regularized} as \emph{inverse transport networks} (ITN).
In practice, our ITNs can be implemented using the encoder-decoder architecture with the decoder replaced with physics-based renderers. A visualization of this architecture is shown in Figure~\ref{fig:architecture}(b).

\boldstart{Training inverse transport networks.}
Using our ITNs requires overcoming a key computational challenge, namely solving the optimization problem~\eqref{eq:erm:regularized} for their training.
The difficulty comes exactly from the light transport operator $\opT$, whose evaluation generally requires solving the radiative transfer~\cite{chandrasekhar1960radiative} and rendering equations~\cite{kajiya1986rendering}.
As computer graphics provides us with rendering algorithms for approximate forward evaluation of $\opT$, training could be performed using algorithms such as REINFORCE~\cite{wuneural}, without differentiating the regularization term. However, such algorithms are known to suffer from slow convergence.

Instead, it would be preferable to optimize the loss~\eqref{eq:erm:regularized} using state-of-the-art stochastic gradient descent algorithms~\cite{sutskever2013importance,kingma2014adam,zeiler2012adadelta,duchi2011adaptive}, also known as backpropagation~\cite{bottou2008tradeoffs,lecun1998gradient,krizhevsky2012imagenet}.
This requires being able to estimate derivatives of the light transport operator $\opT$ with respect to physical parameters $\bpi$ in an unbiased manner, a task we refer to as \emph{differentiable rendering}.
Existing differentiable rendering engines such as OpenDR~\cite{loper2014opendr} can do this computation only for approximate direct lighting models.
This goes against our goal for training networks by accounting for the full light transport process.
Instead, in the next section, we address this computational problem by developing an \emph{efficient}, \emph{general-purpose} and \emph{physically accurate} differentiable rendering engine, based on Monte Carlo integration.

\boldstart{Post-learning refinement.}
At test time, given input images, performing a formward pass through the trained network $\opN$ provides us with a first estimate of the unknown parameters $\bpi$ underlying an input image $I$.
We can further improve the parameter estimate through a second estimation stage, where we use stochastic gradient descent together with Monte Carlo differentiable rendering to minimize the analysis-by-synthesis loss~\eqref{eq:inverse:rendering} for the image $I$. Critically, we can initialize this second estimation stage using the parameter estimate produced by the network.

The effect of this seeding is that the analysis-by-synthesis optimization can converge to a solution much faster than if we had skipped the network-based estimation stage and used a random initial point.
We expect the ITN architecture to be particularly effective for this kind of analysis-by-synthesis acceleration, given that the regularization term in its training loss function~\eqref{eq:erm:regularized} encourages the network to produce estimates that are close to those that would be obtained by directly performing analysis-by-synthesis optimization. 
Additionally, as the network is trained using supervised information, we expect that this initialization will help the optimization procedure converge to the global minimum of the analysis-by-synthesis loss~\eqref{eq:inverse:rendering}, avoiding ambiguities in the physical parameter space.

\boldstart{Relationship to prior work.} Regularization similar to Equation~\eqref{eq:erm:regularized} have previously appeared in two types of literature.
The first is autoencoder architectures~\cite{vincent2010stacked,kingma2013auto} that, in addition to the regressor (\emph{encoder}) network $\opN\bracket{\bw}$ (which maps images to parameters), uses a second \emph{decoder} network $\opD\bracket{\bu}$ that maps the parameters back to images.
Then, the regularization term in Equation~\eqref{eq:erm:regularized} is replaced with $\norm{I_d - \opD\bracket{\bu}\paren{\opN\bracket{\bw}\paren{I_d}}}^2$, and both the encoder and decoder networks are trained simultaneously.
These architectures are of great utility when seeking to infer semantic parameters (e.g., a class label) about a scene, in which case there is often no analytical model for the forward mapping of these parameters to images.
However, when the unknowns $\bpi$ are physical parameters, autoencoder architextures do not take advantage of the rich knowledge we have from the physics governing the forward operator $\opT$.
Additionally, the learned forward mapping $\opD\bracket{\bu}$ may not generalize to unseen instances, as it is specific to the data set used to train the autoencoder. Figure~\ref{fig:architecture} compares the autoencoder and inverse transport architectures.

The second type of architectures are networks using regularization terms with the light transport operator $\opT$ approximated by some  $\opS$~\cite{lime,shrenderer,liu2017material,mofa,neuralmeshrenderer}.
These approximations are generally based on direct lighting models for image formation, where photons are assumed to only interact with the scene once between leaving a light source and arriving at a detector (e.g., direct reflection without interreflections, single scattering).
Consequently, these networks perform suboptimally in cases where higher-order transport effects are predominant.
Inspired by these prior works, our ITNs are general physics-aware learning pipelines that can be used for solving inverse problems involving light transport of arbitrary complexity.
To demonstrate the importance of using the full light transport operator $\opT$ instead of direct-lighting approximations $\opS$, we compare in Section~\ref{sec:application} these approaches as well as the unregularized approach of Equation~\eqref{eq:erm}, for the problem of inferring scattering parameters of translucent materials.
We choose this application as an instance of an extreme multi-path, multi-bounce transport problem, where single scattering approximations have limited applicability.

\section{Differentiable Monte Carlo Rendering}
\label{sec:dr}

\boldstart{Background.} We aim to develop algorithms for estimating derivatives of radiometric quantities, e.g., intensity measured by a detector, with respect to physical scene properties, e.g., optical material of objects in the scene, in an unbiased manner. Our formulation directly borrows from the path integral formulation for forward rendering. Therefore to make our discussion self-contained, we provide here the necessary background. Our starting point is the expression of radiometric quantities as integrals over the space of possible light particle paths~\cite{veach1998robust}:
\begin{equation}
\label{eq:path}
\opT\paren{\bpi} = \int_{\Path} \glthrough\bracket{\bpi}\paren{\path} \,\mathrm{d} \path,
\end{equation}
where, for any $K > 1$, $\path := (\bx_0, \bx_1, \ldots, \bx_K)$ with $\bx_i \in \R^3$ (for $i = 0, 1, \ldots, K$) indicates a light transport path with $\bx_0$ located on a light source and $\bx_K$ on a sensor, and
\begin{equation}
\begin{aligned}
\glthrough\paren{\path} =\;& G\bracket{\bpi}\paren{\bx_{K - 1}, \bx_K}\\
&\prod_{k = 1}^{K-1} G\bracket{\bpi}\paren{\bx_{k - 1}, \bx_k} \through\bracket{\bpi}(\bx_{k - 1}, \bx_k, \bx_{k + 1}).
\end{aligned}
\end{equation}

\begin{figure*}[t]
	\centering
	\includegraphics[width=\textwidth]{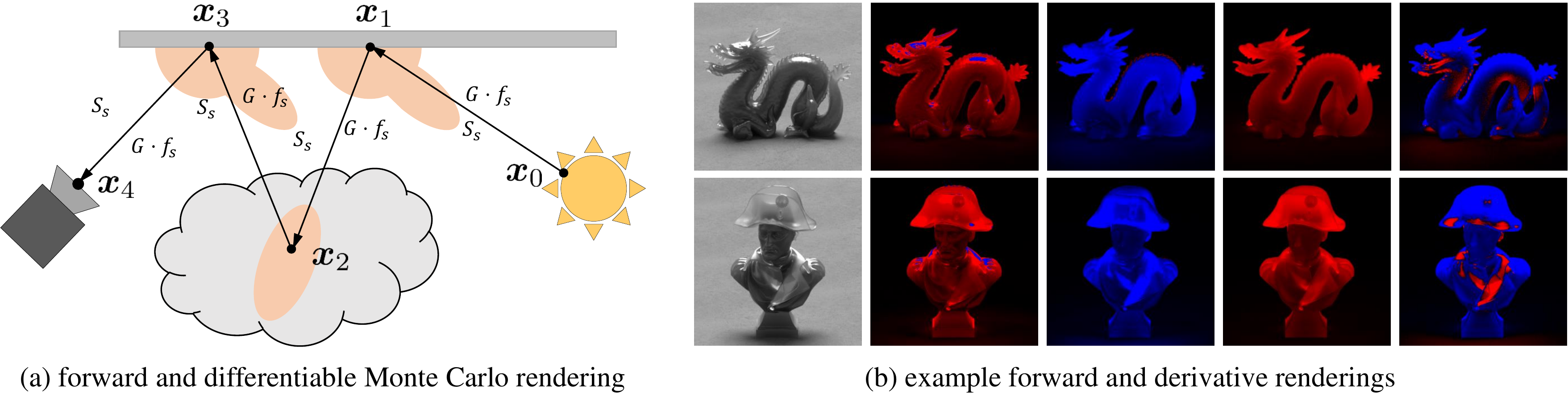}
	\caption{\label{fig:dr}
		\textbf{Monte Carlo forward and differentiable rendering:} (a) Monte Carlo rendering estimates radiometric measurements by randomly sampling photon paths $\path$ and aggregating their radiometric throughput $\glthrough$. By additionally evaluating and aggregating the radiometric score $\glscore$ for each path, we can use the same procedure to additionally estimate \emph{derivatives} of radiometric measurements with respect to physical scene parameters. (b) Example forward and differentiable renderings for two different scenes. In each scene, the object's surface reflectance and transmittance are characterized by a rough dielectric BSDF with roughness parameter $r$, and its subsurface scattering is characterized by spatially-homogeneous extinction coefficient $\sigma_t$, volumetric albedo $\albedo$, and Henyey-Greenstein phase function with some value $g$. For each of the two objects, we show from left to right: a forward rendering, and rendered derivatives with respect to $r$, $\sigma_t$, $\albedo$, and $g$. In the differentiable renderings, red indicates positive and blue negative values. Additionally, most of the background in the differentiable renderings is black, as the intensities at those parts of the image are largely independent of the object's material parameters.}
\end{figure*}

This integration is performed over the space 
$\Path$ of all possible paths. 
In each such path, 
the intermediate points $\bx_k$ with $0 < k < K$ capture light-scene interactions via reflection, refraction, and subsurface scattering.
The \emph{throughput function} $\glthrough\bracket{\bpi}$ describes the amount of radiance contributed by a path as a function of the scene geometry, material properties, illumination and detector characteristics.
The throughput can be expressed as the product of per-vertex terms $G\bracket{\bpi}\through\bracket{\bpi}$, whose exact role varies for different points $\bx_k$.
In particular, when $\bx_k$ is a point on the surface of an object, then $\through\bracket{\bpi}$ is equal to the bidirectional scattering distribution function (BSDF) of the object at point $\bx_k$ with normal $n\paren{\bx_k}$.
If one of $\bx_{k-1}$ and $\bx_{k+1}$ is outside the object and the other inside, then $\through\bracket{\bpi}$ describes a refraction event; otherwise it describes a reflection event. If $\bx_{k}$ is a point inside a scattering medium, then $\through\bracket{\bpi}$ describes a scattering event and is equal to the product of the medium's volumetric albedo and phase function at $\bx_{k}$. In both cases, the BSDF and phase function are evaluated on incoming and outgoing directions $\bx_{k-1}\rightarrow\bx_k$ and $\bx_{k}\rightarrow\bx_{k+1}$, respectively. Finally, the term $G\bracket{\bpi}$ equals binary visibility when both $\bx_{i - 1}$ and $\bx_i$ are outside scattering media. Otherwise, it equals volumetric attenuation, and is a function of the medium's extinction coefficient. A visualization of this is shown in Figure~\ref{fig:dr}(a).

The path integral formulation of Equation~\eqref{eq:path} accurately describes light transport for scenes of arbitrary complexity, including higher-order transport effects such as interreflections and multiple scattering that cannot be represented using direct lighting approximations.
Unfortunately, except for trivial scenes, Equation~\eqref{eq:path} cannot be evaluated analytically.
Computer graphics has focused on the approximation of Equation~\eqref{eq:path} using Monte Carlo integration~\cite{AGIbook2,veach1998robust,pharr2016physically}, which at a high-level operates as follows.
First, a set of paths $\{\path_n: n=1,\dots,N\}$ are sampled stochastically from a probability density $p$ defined on the path space $\Path$. Second, the throughput $\through\bracket{\bpi}$ of each these paths is computed. Third and final, an unbiased and consistent estimator of Equation~\eqref{eq:path} is formed as:
\begin{equation}
\langle \opT\paren{\bpi} \rangle = \frac{1}{N} \sum_{n=1}^N \frac{\glthrough\bracket{\bpi}\paren{\path_n}}{p\paren{\path_n}}.\label{eq:path_MC}
\end{equation}
The performance of these estimators depends critically on the propability distribution $p$ used to sample light transport paths. Modern Monte Carlo rendering techniques use path sampling techniques such as path tracing~\cite{Kajiya:1986:RE}, bidirectional path tracing (BDPT)~\cite{lafortune1996mathematical}, and Metropolis light transport (MLT)~\cite{veach1998robust}, which simulate path sampling densities $p$ similar to the throughput function $\through\bracket{\bpi}$ in order to reduce the variance of the estimator~\eqref{eq:path_MC}.

\subsection{Differentiable rendering}

Instead of image measurements $\opT\paren{\bpi}$, we focus on estimating their derivatives $\partial \opT\paren{\bpi}/\partial \bpi$ with respect to scene parameters $\bpi$ describing illumination, geometry,  optical material.
To this end, we follow previous work~\cite{Gkioulekas2013:IVR,Zhao:2016:DSP,Khungurn:2015:MRF,gkioulekas2016inverse} that generalizes the  path integral formulation to apply for such derivatives.
In this setting, differentiating Equation~\eqref{eq:path} and rearranging the throughput terms yields:
\begin{align}
&\partial \opT\paren{\bpi}/\partial \bpi = \int_{\Path} \glthrough\bracket{\bpi}\paren{\path}\glscore\bracket{\bpi}\paren{\path} \,\mathrm{d} \path,\label{eq:path_diff} \\
&\text{where} \quad\glscore\bracket{\bpi}\paren{\path} = 
\sum_{k=1}^{K-1}\score\bracket{\bpi}\paren{\bx_{k-1},\bx_k,\bx_{k+1}},\label{eq:gradient:sum} \\
&\score\bracket{\bpi}\paren{\bx_{k-1},\bx_k,\bx_{k+1}} =
\frac{\partial \through\bracket{\bpi}\paren{\bx_{k-1},\bx_k,\bx_{k+1}} / 
\partial \bpi}{\through\bracket{\bpi}\paren{\bx_{k-1},\bx_k,\bx_{k+1}}}. 
\label{eq:segment:score}
\end{align}
Compared to Equation~\eqref{eq:path}, the path integral for the derivative case includes the \emph{score function} $\glscore\bracket{\bpi}$, that sums derivatives of the per-vertex throughput with respect to scene parameters $\bpi$. Similar to the forward rendering case, we can combine this path integral expression with Monte Carlo integration, in order to form a consistent and unbiased estimate of the derivative as:
\begin{align}
\label{eq:path_diff_MC}
\langle \partial \opT\paren{\bpi}/\partial \bpi \rangle &= \frac{1}{N}\sum_{n=1}^N \frac{\glthrough\bracket{\bpi}\paren{\path_n}\glscore\bracket{\bpi}\paren{\path_n}}{p\paren{\path_n}}.
\end{align}
When implementing Monte Carlo differentiable rendering, an important challenge is computing the derivative terms involved in the evaluation of the score function $\glscore\bracket{\bpi}$. In several cases, it is possible to derive analytical expressions for the derivatives of the throughput function $\glthrough\bracket{\bpi}$ with respect to certain scene parameters $\bpi$. This approach has been used previously to estimate derivatives with respect to scattering material parameters~\cite{Gkioulekas2013:IVR,Zhao:2016:DSP,Khungurn:2015:MRF,gkioulekas2016inverse}.

Here we take a different approach: Instead of hard-coding derivatives for a pre-defined set of parameters, we combine a general-purpose Monte Carlo renderer with automatic differentiation~\cite{griewank1989automatic,rall1981automatic}.
This allows us to compute physically accurate derivatives of the throughput function $\glthrough\bracket{\bpi}$ with respect to arbitrary scene parameters.

\boldstart{Implementation.}
We have developed a new simulation engine named \emph{{\mtsadname}} that integrates the Stan Math Library~\cite{carpenter2015stan} for automatic differentiation, with the Mitsuba engine~\cite{mitsuba} for physically accurate Monte Carlo rendering.
{\mtsadname} supports the same variety of materials, light sources, and camera models as the original Mitsuba renderer, and currently supports differentiation with respect to the following types of scene parameters:
\begin{itemize}
\item Spatially constant and varying BSDF models, including ideal Lambertian and specular reflectance, physics-inspired microfacet models~\cite{cook1982reflectance} for rough diffuse~\cite{oren1995generalization} and specular~\cite{walter2007microfacet,ward1992measuring,ashikhmin2000anisotropic} reflectance and refraction, as well as dictionary BSDF representations~\cite{matusik2003data}. For each of these models, differentiation is possible with respect to their albedo, roughness, index of refraction, or mixing weight parameters, as applicable. 
\item Spatially constant and varying scattering, including a variety of phase function models such as Henyey-Greenstein~\cite{henyey1941diffuse}, von-Mises Fisher~\cite{gkioulekas2013understanding}, and their linear combinations. Differentiation is possible with respect to the extinction coefficient, single-scattering albedo, and phase function parameters.
\item Geometry, such as spatially varying bump maps and normal maps that can be differentiated with respect to depth and normal displacements, respectively.
\item Environment map illumination, including the Hošek-Wilkie sky and sun-sky models~\cite{hovsekhovsek2013adding,hosek2012analytic}, and dictionary representations such as spherical harmonics. 
\end{itemize}
Figure~\ref{fig:dr}{b} shows examples of rendered image derivatives with respect to a few of these parameters. Additionally, {\mtsadname} is designed to be easily extensible for differentiation with respect to other scene properties. Additional capabilities, such as new BSDF models, can be incorporated using the plugin system of Mitsuba.

\boldstart{Path sampling algorithms.}
Although the differentiated estimator~\eqref{eq:path_diff_MC} has a similar form to the original one~\eqref{eq:path_MC}, the difference between $\glthrough\bracket{\bpi}$ and $\glthrough\bracket{\bpi}\glscore\bracket{\bpi}$ suggests that efficient derivative estimation would require the development of new path sampling techniques that generate paths from probability distributions $p$ that approximate $\glthrough\bracket{\bpi}\glscore\bracket{\bpi}$. However, it has been observed empirically that derivative estimation can still be done efficiently using the same path sampling techniques as for forward rendering~\cite{Zhao:2016:DSP,Khungurn:2015:MRF,gkioulekas2016inverse}.
{\mtsadname} follows this approach, and uses path tracing for both forward and differentiable rendering.

\boldstart{Stochastic optimization.}
In addition to its generality and physical accuracy, Monte Carlo differentiable rendering provides computational advantages in the context of gradient-based optimization.
In particular, learning deep neural networks from large training datasets using empirical risk minimization, as in Equation~\eqref{eq:erm}, greatly relies on the ability to perform backpropagation in a \emph{stochastic} manner.
Namely, one can compute derivatives of the loss function using stochastic subsets of the training set (\emph{minibatches}).
Changing the size of the minibatches allows controlling the tradeoff between the cost of gradient computations and the number of iterations until convergence~\cite{bottou2008tradeoffs,lecun1998gradient,krizhevsky2012imagenet}.

Monte Carlo differentiable rendering offers control over a similar capability: we can reduce the number of sampled paths to speedup derivative computation at the cost of increased variance.
As the derivative estimations~\eqref{eq:path_diff_MC} remain consistent and unbiased, we can use this to take advantage of the same convergence guarantees and tradeoffs as with stochastic backpropagation.
Therefore, our Monte Carlo differentiable rendering engine is particularly well-suited for training of neural networks using state-of-the-art stochastic gradient descent algorithms~\cite{sutskever2013importance,kingma2014adam,zeiler2012adadelta,duchi2011adaptive}.

\section{Application: Inverse Scattering}\label{sec:application}
To demonstrate the applicability and utility of our proposed ITN architecture, we focus on the problem of \emph{homogeneous inverse scattering}: given an image of a translucent object of known shape under known illumination, we aim to determine the optical material parameters that control the scattering of light inside this object. Specifically, the material is characterized by a triplet of macroscopic bulk parameters $\bpi = \curly{\ext, \albedo, \phase}$ as follows.
\begin{itemize}
	\item The \emph{extinction coefficient} $\ext$ is a scalar corresponding to the optical density of the material and controls the average distance between consecutive volume events (also known as \emph{mean free path}).
	\item The \emph{volumetric albedo} $\albedo$ is a scalar probability and controls whether a photon is scattered or absorbed at a volume event.
	\item The \emph{phase function} $\phase$ is a probability distribution over the sphere of directions and controls the direction scattered photons continue to travel towards. 
\end{itemize}
In general, these parameters can be spatially varying, but in our application we assume them to remain constant everywhere inside the object (homogeneous scattering).

Inverse scattering is a problem that is particularly well-suited for ITNs because of the characteristic complexity of light transport within translucent objects.
Each photon propagating inside a scattering medium undergoes a random walk, controlled non-linearly by the medium's bulk parameters. These random walks typically involve more than one bounces.
In turn, a radiometric detector capturing an image of such an object accumulates a large number of photons, each performing a different random walk.
Therefore, images of translucent objects are a typical example involving extremely multi-path and multi-bounce light transport.
Although it is possible to simplify this image formation process by assuming that each photon only bounces once inside the object, this \emph{single-scattering} approximation is only applicable to very optically thin materials such as diluted liquids~\cite{narasimhan2006acquiring}, necessitating the development of general-purpose inverse scattering techniques to accurately model the full complexity of volumetric light transport~\cite{Gkioulekas2013:IVR}.

\boldstart{Loss functions and architectures.}
We use the homogeneous inverse scattering setting to compare neural networks trained with three different loss functions.
The first network is trained using the purely supervised loss~\eqref{eq:erm}.
We refer to this as the \emph{regressor network} (RN).
The second network is trained using the regularized loss~\eqref{eq:erm:regularized}, where the $\opT$ corresponds to simulating the full volumetric light transport.
This is the \emph{inverse-transport network} (ITN) introduced in Section~\ref{ssec:itn}.
The third network is trained again using the regularized loss~\eqref{eq:erm:regularized}, but this time with $\opT$ replaced with the single-scattering approximation $\opS$ to volumetric light transport.
We term this the \emph{single-scattering network} (SSN).

For all the three networks, we use the structure proposed by Liu et al.~\cite{liu2017material}.
Specifically, each network is composed of seven convolutional layers, and the size of the output channel for each layer is reduced to half the size of its input.
Each convolutional layer is followed by a rectified linear unit (ReLU) and a max-pooling layer. A fully connected layer follows the convolutional layers at the end.

For both the ITN and SSN, we perform forward and differential evaluations of the transport operators $\opT$ and $\opS$, respectively, using {\mtsadname}.
The rendering layer is connected to the network and takes as input the material parameters $\bpi=\curly{\ext, \albedo, \phase}$ predicted by the network, as shown in Figure~\ref{fig:architecture}(b). These parameters are used to render images and their per-parameter derivatives, as needed for evaluating the loss function and performing backpropagation.

\begin{figure*}[t]
	\centering
	\includegraphics[width=\textwidth]{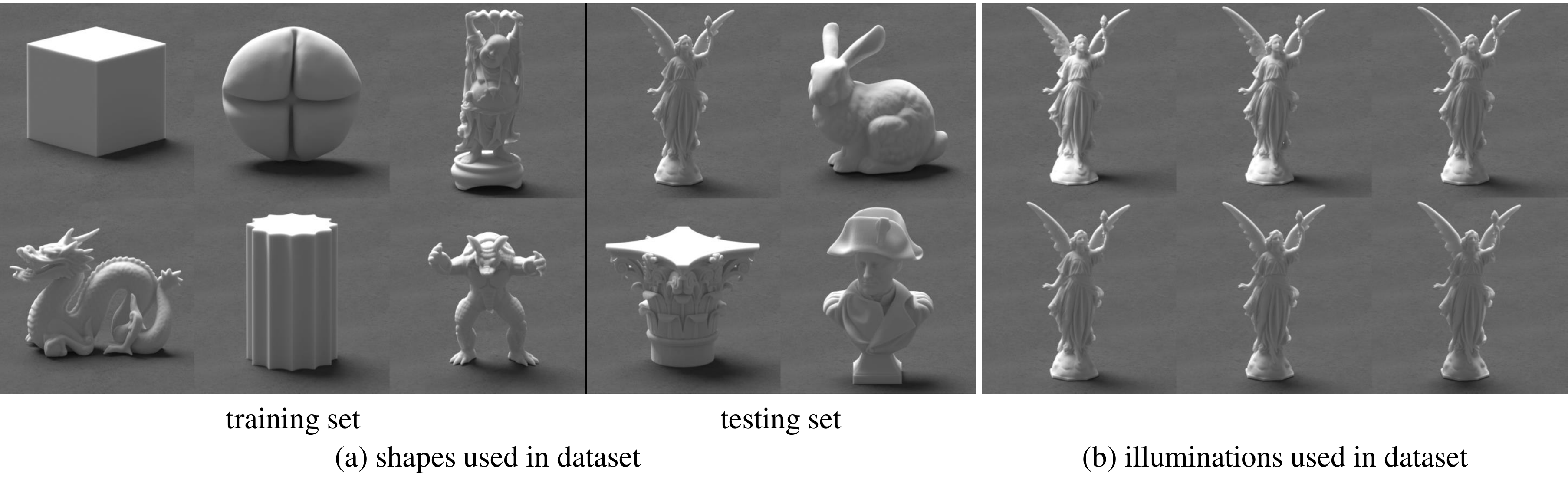}
	\caption{\label{fig:dataset}
		\textbf{Dataset for homogeneous inverse scattering:} We render 20,000 images by combining different shape, illumination, and scattering material conditions. (a) For shape, we use ten meshes commonly encountered in the computer vision and graphics literature. We split these into shapes used only for training, and shapes used only for testing. (b) For illumination, we use the Ho\v{s}ek-Wilkie sun-sky model under different orientations, corresponding to several illumination directions between side-lighting and back-lighting.}
\end{figure*} 

\boldstart{Datasets and evaluation.}
For our quantitative comparisons, we use a synthetic dataset containing images of translucent objects with varying geometry, illumination, and scattering parameters.
We use ten different object shapes, as shown in Figure~\ref{fig:dataset}, selected among common computer graphics meshes. Each shape is placed under ten different illumination conditions created using the Ho\v{s}ek-Wilkie sun-sky model~\cite{hovsekhovsek2013adding,hosek2012analytic}.
Finally, for each shape and illumination combination, we render images for different scattering parameter triplets $\bpi$, spanning the following ranges for each individual parameter: $\ext\in\bracket{\unit[25]{\text{mm}^{-1}},\unit[300]{\text{mm}^{-1}}}$, $\albedo\in\bracket{0.3,0.95}$, and Henyey-Greenstein phase functions $\phase$ with parameter $g\in\bracket{0,0.9}$.
This covers materials with translucent appearance ranging from near-transparent to near-opaque. We use the Mitsuba physics-based renderer~\cite{mitsuba} to simulate $20,000$ high-dynamic range images under these settings.

We use the images corresponding to six shapes and four illuminations as the training set, and use all of the remaining images for testing.
This yields a testing set that includes images of objects of unseen shape, or under unseen illumination, or both.
We use this separation between training and testing images specifically in order to evaluate the generalization properties of the three different neural network architectures.
Specifically, we compare the RN, ITN and SSN in terms of how accurately they can predict material parameters, how accurately they can reproduce input image appearance, and how much they can accelerate analysis-by-synthesis, for images from both the training and test sets.

\section{Experiments}\label{sec:experiments}

When training the ITN and SSN, we use as initialization a network trained with only the supervised loss of Equation~\eqref{eq:erm:regularized} for few epochs. We set $\lambda$ in Equation~\eqref{eq:erm:regularized} so that the supervised loss and regularization term for both ITN and SSN have approximately the same magnitude. All networks are trained using the Adam optimization algorithm~\cite{kingma2014adam}. 

We evaluate the three networks, RN, ITN and SSN, in three ways: First, we consider how accurately they predict the material parameters unrelying input images. Second, we examine how well images rendered with the predicted parameters match the appearance of the corresponding input images. Third, we evaluate the utility of parameter predictions for initializing inverse rendering optimization problems of the form of Equation~\eqref{eq:inverse:rendering}.	

\begin{table}[ht]
    \centering
    \begin{tabularx}{\linewidth}{X|ccc|ccc}
\multirow{2}{*}{network} & \multicolumn{3}{c|}{training} & \multicolumn{3}{c}{testing} \\ \cline{2-7}
           & $\ext$ & $\albedo$ & $g$ & $\ext$ & $\albedo$ & $g$\\ \hline
RN & \textbf{24.43} & \textbf{0.06} & \textbf{0.13} & 81.59 & 0.15 & 0.41 \\
ITN & 41.80 & 0.08 & 0.25 & \textbf{57.44} & \textbf{0.10} & \textbf{0.25} \\
SSN & 56.50 & 0.17 & 0.34 & 65.06 & 0.18 & 0.35
\end{tabularx}

    \caption{Average RMSE of parameters predicted using RN, ITN and SSN.}
    \label{table:paramserror}
\end{table}

\boldstart{Parameter prediction.} For this evaluation, we use RN, ITN and SSN to predict three material parameters $\bpi=\curly{\ext, \albedo, g}$ for input images coming from both the training and testing datasets. We compare them with ground-truth values and present the average root-mean-square error (RMSE) in Table~\ref{table:paramserror}. We observe that, even though the RN can achieve lower prediction error on the training set, the ITN performs better on the testing set. These results provides evidence that the regularization term in the Equation~\eqref{eq:erm:regularized} used for training the ITN allows the network to generalize better to new scenes with unsheen shapes and illumination. Additionally, by considering the performance of the SSN on the testing set, we observe that it is not sufficient to use the single-scattering operator $\opS$ (as in SSN). Instead, better generalization requires using the full volumetric light transport operator $\opT$ (as in ITN).

\begin{table}[ht]
    \centering
    \begin{tabularx}{\linewidth}{X|cc|cc}
net- & \multicolumn{2}{c|}{training} & \multicolumn{2}{c}{testing} \\ \cline{2-5}
work &RMSE & 1-MS-SSIM & RMSE & 1-MS-SSIM \\ \hline
RN & 0.3368 & 0.0665 & 0.3180 & 0.1002 \\
ITN & \textbf{0.3299} & \textbf{0.0609} & \textbf{0.1631} & \textbf{0.0426} \\
SSN & 0.9444 & 0.1673 & 0.8735 & 0.1585 \\
\end{tabularx}

    \caption{Average RMSE and $1 - \text{MS-SSIM}$ of image appearance predicted using RN, ITN and SSN.}
    \label{table:appearanceserror}
\end{table}

\boldstart{Appearance reproduction.} To evaluate the networks in terms of their ability to reproduce input image appearance, we use the material parameters predicted by the RN, ITN and SSN to
render images with the same shape and illumination settings as what is used for the input image. We then compare the rendered and input images using two different metrics: (1) RMSE, and
(2) the multi-scale structural similarity index (MS-SSIM)~\cite{wang2003multiscale}, as a representative perceptually-motivated image quality metric. We average each of these metrics across both the training and testing set, as shown in Table~\ref{table:appearanceserror}. (Note that we report $1 - \text{MS-SSIM}$ instead of MS-SSIM values, to ensure that across all metrics lower values indicate better performance.)

We note that, unlike the case of parameter prediction (Table~\ref{table:paramserror}), here the ITN outperforms the RN in terms of appearance prediction on both the training and testing set. This improved performance is particularly pronounced in the case of the unseen scenes in the testing set, where the ITN results in an improvement in appearance prediction of about 50$\%$. In contrast, SSN has the highest appearance errors in both training and testing sets. This suggests that the single-scattering approximation employed by the SSN is not sufficient for accurately reproducing input images whose appearance is the result of higher-order light transport. 

\begin{table}[ht]
    \centering
    \addtolength{\tabcolsep}{-2.5pt}
\begin{tabularx}{\linewidth}{Xl|ccccc}
\multirow{2}{*}{} &\multirow{2}{*}{}& \multicolumn{5}{c}{number of iterations}\\ \cline{3-7}
                       && 1 & 50 & 100 & 150 & 200  \\ \hline
\multirow{2}{*}{loss}   & RN   & 0.0670 & 0.0041 & 0.0023 & 0.0018 & 0.0015 \\
                      & ITN & \textbf{0.0189} & \textbf{0.0024} & \textbf{0.0019} & \textbf{0.0015} & \textbf{0.0013} \\ \hline
\multirow{2}{*}{$\log(\ext)$}& RN & \textbf{0.1506} & \textbf{0.0488} & 0.0346 & 0.0265 & 0.0224\\
                      & ITN &  0.2084 & 0.0805 & \textbf{0.0191} & \textbf{0.0090} & \textbf{0.0047} \\ \hline                      
\multirow{2}{*}{$\albedo$} & RN & 0.0092 & \textbf{0.0009} & \textbf{0.0006} & 0.0006 & 0.0005\\
                      & ITN &  \textbf{0.0039} & 0.0010 & 0.0008 & \textbf{0.0003} & \textbf{0.0002} \\ \hline
\multirow{2}{*}{$g$} & RN & 0.0812 & 0.0404 & 0.0205 & 0.0121 & 0.0069 \\
                      & ITN & \textbf{0.0354} & \textbf{0.0276} & \textbf{0.0143} & \textbf{0.0065} & \textbf{0.0031}  \\
\end{tabularx}

    \caption{Medians MSE of image loss, $\log\paren{\ext}$ , $\albedo$ and $g$ for both RN and ITN}
    \label{table:inverseerror}
\end{table}

\boldstart{Initialization of inverse rendering.} Finally, we compare the parameter predictions of the RN and ITN in terms of how much they accelerate analysis-by-synthesis when used to initialize optimization problems of the form of Equation~\eqref{eq:inverse:rendering}. For this, we randomly select 10 images rendered with each shape in the test set, and use them to perform inverse rendering. We use the parameters predicted by RN and ITN to initialize the inverse rendering process, which is run for a fixed number of 200 stochastic gradient descent iterations. Every 50 iterations (including at the initialization), and we record the median
value of the appearance loss used for the optimization (Equation~\eqref{eq:inverse:rendering}), as well as the errors between the groundtruth parameters and the parameters at the end of that iteration. 

From Table~\ref{table:inverseerror}, we observe that the parameter predictions produced by the ITN are more useful for initialization, as they result in lower values of the loss function after the same number of iterations. Additionally, we observe that the parameters the inverse rendering procedure converges to are closer to the groundtruth when using ITN-based initialization: Before any optimization iterations, all three parameters are initialized to be closer to the ground-truth value except for $\log\paren{\ext}$. After a fixed number of iterations, the median parameter errors of the procedures initialized using ITN are uniformly smaller than those initialized using RN.

\section{Conclusions}\label{sec:conclusions}

We introduced inverse transport networks as a new neural network architecture that can be used in inverse problems where it is necessary to predict \emph{physical} parameters (shape, material, illumination) underlying some input images. These networks are trained so that their parameter predictions not only approximate groundtruth parameters (supervised loss), but also can be used to synthesize images that closely match the corresponding input images (unsupervised regularization). Our experiments show that, when image synthesis is performed using physically-accurate rendering algorithms that capture all light transport effects in the input images, this regularization significantly improves the generalization of trained networks to inputs consisting of previously unseen shapes, illumination, or both.

This improved generalization performance comes at a high computational cost, due to the need to minimize a training function that includes forward Monte Carlo rendering operations. We alleviated this cost by introducing a general-purpose, physically-accurate \emph{differentiable renderer}. This renderer allows us to estimate derivatives of images with respect to physical scene parameters, which in turn means that we can use efficient stochastic gradient descent procedures to train the inverse transport networks.

By demonstrating the utility of inverse transport networks for physical inference tasks, and of differentiable renderers for training such networks, we hope to motivate future work in both computer graphics and computer vision. In computer graphics, an exciting direction of research is the development of optimal path sampling techniques for differentiable rendering, which so far has received limited attention~\cite{Gkioulekas2013:IVR}. In computer vision, the development of differentiable renderers opens up a whole new direction of exploration, as researchers investigate more general learning architectures that intelligently combine neural networks with physics-based simulation.

\bibliographystyle{ieee}
\bibliography{gfxmlinverse,kb}
\end{document}